\pdfoutput=1
\documentclass[11pt]{article}

\usepackage[final]{acl}

\usepackage{times}
\usepackage{latexsym}
\usepackage[normalem]{ulem}
\usepackage[T1]{fontenc}
\usepackage[utf8]{inputenc}

\usepackage{microtype}

\usepackage{inconsolata}

\usepackage{graphicx}

\usepackage{enumitem}

\title{More Data, Fewer Diacritics: Scaling Arabic TTS}

\author{Ahmed Musleh, Yifan Zhang, Kareem Darwish \\
  QCRI, HBKU, Qatar \\
  \texttt{\{ahmusleh,yzhang,kadarwish\}@hbku.edu.qa} \\
  }

\begin{document}
\maketitle
\begin{abstract}
%\sout{Arabic TTS research has been stinted by the limited amount of publicly available training data and the availability of accurate Arabic diacritization models.}
Arabic Text-to-Speech (TTS) research has been hindered by the  availability of both publicly available training data and accurate Arabic diacritization models. In this paper, %\sout{we address both limitations}
we address the limitation by exploring Arabic TTS training on  large automatically annotated data.  Namely, we built a robust pipeline for collecting Arabic recordings and processing them automatically using voice activity detection, speech recognition, automatic diacritization, and noise filtering, resulting in around 4,000 hours of Arabic TTS training data.  We then trained several robust TTS models with voice cloning using varying amounts of data, namely 100, 1,000, and 4,000 hours with and without diacritization.  We show that though models trained on diacritized data are generally better, larger amounts of training data compensate for the lack of diacritics to a significant degree.  We plan to release a public Arabic TTS model that works without the need for diacritization.
\end{abstract}

\section{Introduction}
% \sout{Arabic Text-to-Speech (TTS) synthesis, with the overarching objectives of achieving highly intelligible and natural output, is crucial for numerous applications such as pedagogical applications and voice agents.}
Arabic Text-to-Speech (TTS) synthesis plays a vital role in enabling speech interfaces -- the most natural and intuitive mode of human-computer interaction. Achieving highly intelligible and natural-sounding Arabic TTS is therefore essential for a wide range of applications, including educational tools and voice-based virtual assistants. While substantial progress has been observed in TTS development for many languages, particularly English and Chinese, Arabic continues to lag behind and faces unique challenges stemming from the limited availability of large high-quality training datasets and the need for accurate text diacritization.  Diacritics, which are phonetic markers associated with letters, are routinely omitted in standard written texts, introducing much ambiguity for automated computational systems.   This paper presents a comprehensive approach to advancing state-of-the-art Arabic TTS, focusing on data scale, minimizing the need for diacritization.

Our first major contribution is the development of a robust, automated pipeline for curating accurate Arabic audio and text (diacritized and undiacritized). This pipeline leverages ever improving technologies including Voice Activity Detection (VAD), Automatic Speech Recognition (ASR), automatic diacritization, and noise filtering, to efficiently curate a massive dataset. With our proposed pipeline, we successfully assembled a large Arabic TTS training corpus composed of roughly 4,000 hours.  Although these technologies are imperfect, we show that the errors they produce become less and less consequential as the collected data grows leading to improved TTS performance. % Leveraging our proposed pipeline we were able to collect more than 4,000 hours of high quality audio with associated transcripts with and without diacritics. 
% Arabic Diacritics, while significant in helping TTS systems, is routinely omitted from written texts, leaving users who want a more accurate result from TTS systems having to figure out a way to diacritize their text.

Our second contribution involves training and evaluating several neural TTS models, with and without diacritization, capable of high-quality voice cloning with minimal enrollment audio (demonstrated effectively with as little as 3 seconds). We systematically investigate the impact of training data volume and the presence of explicit diacritization, training models on datasets of varying sizes, namely: 100, 1,000, and 4,000 hours.    %, both with and without the automatically derived diacritics.

Third, we provide insights into the data requirements for Arabic TTS. While models trained on text inputs with explicit diacritization generally achieve superior synthesis quality compared to their non-diacritized counterparts, our experiments clearly demonstrate that this performance gap diminishes as the total volume of training data increases. This suggests that large-scale non-diacritized data can compensate, to a significant extent, for the lack of explicit diacritic information.  SOTA diacritizers are trained with more than 5 million word diacritized corpora, which is equivalent to the speech spoken in roughly 500 to 625 hours of audio (assuming that each audio hour has 8-10k words).  Thus, theoretically, a TTS model trained on more than that amount can potentially learn effective diacritization implicitly.

Finally, based on these findings and to foster further research and application development in Arabic TTS, we release a high-quality, publicly available Arabic TTS model. A core feature of this released model is its capability to synthesize speech effectively from raw, non-diacritized Arabic text inputs, simplifying its practical deployment. Our work provides valuable resources, experimental insights, and a performant public model for the Arabic speech community.

\section{Related Work}
\subsection{TTS Modeling}
The field of text-to-speech (TTS) synthesis is rapidly advancing, with a strong focus on developing efficient, non-autoregressive generative models that deliver high-fidelity, zero-shot performance. \citet{eskimez2024e2} introduced E2-TTS, an ``embarrassingly easy'' paradigm using a non-autoregressive, flow-matching architecture to achieve state-of-the-art results with a simplified pipeline. F5-TTS builds upon this framework, specifically addressing the training instability and robustness challenges of the original model by integrating a ConvNeXt architecture \cite{chen-etal-2024-f5tts}.
This approach aligns with a broader trend of leveraging Diffusion Transformers (DiT) for speech synthesis \cite{popov2021grad} with the 5 best performing non-proprietary models on TTS Arena \cite{tts-arena}, such as Kokoro\footnote{\url{https://huggingface.co/hexgrad/Kokoro-82M}} Fish Speech \cite{fish-speech-v1.4}, and MetaVoice \footnote{\url{https://github.com/metavoiceio/metavoice-src}}, being DiT based \cite{fish-speech-v1.4,li2023styletts}. These models underscore a collective move towards simpler, end-to-end systems that eliminate complex intermediate representations.  In this paper, we use F5-TTS to train our Arabic TTS systems.

%For instance, DiTTo-TTS \cite{lee2025ditto} demonstrated the scalability of DiT-based models, achieving high performance without traditional domain-specific components like phoneme predictors. Similarly, DiTAR \cite{jia2025ditar} introduced a novel patch-based autoregressive method combining a DiT with a language model to reduce computational load. These models underscore a collective move towards simpler, end-to-end systems that eliminate complex intermediate representations.
% The pursuit of efficiency is a central theme in recent work. Flow matching, a core component of our model, is prized for its speed, and the field’s velocity is such that subsequent research has already demonstrated a 9x inference speed-up for F5-TTS by eliminating Classifier-Free Guidance (Liang et al., 2025). Within this dynamic landscape, F5-TTS contributes by refining a promising new architecture, enhancing its robustness and practicality while advancing the collective push towards simpler, faster, and more scalable TTS systems.
\vspace{-6pt}
\subsection{Arabic TTS Data}
Early concatenative \cite{hunt1996unit} and HMM \cite{zen2007hmm} based speech synthesis models required curation of phonetically segmented and aligned datasets, requiring considerable time and effort \cite{yuan2013automatic}. The publicly available Arabic TTS datasets with such segmentation and alignment include a dataset composed of 4 hours of audio developed by \citet{halabi2016arabic} and another containing 7 hours created by \citet{almosallam2013sassc}. % Manual phonetic segmentation and alignment was costly and many approaches were taken to automate this process \cite{yuan2013automatic}. Even with automating alignment and segmentation, there is still the manual verification process where a team of linguists would be present to post-check the alignment. 
The aforementioned DiT models %Tacotron  
alleviate the need for such a dataset and opens the door for training on larger amounts of audio with corresponding transcriptions, aligned at sentence level. \citet{abdelali2022natiq} trained two TACOTRON 2 based Arabic TTS models \cite{wang2017tacotron} using roughly 7 and 10 hours by a female voice actress and male voice actor respectively.  Their dataset is not publicly available. ClArTTS \cite{kulkarni2023clartts} created a 12-hour single speaker corpus for classical and modern standard Arabic. More recently, \citet{olamide2025arvoice} have introduced a substantially larger 83-hour hybrid dataset including 10 hours voice recording and 73 hours synthesized data. To date, this is still the largest publicly available dataset.  All text in these datasets is fully diacritized to match the recorded speech.  \citet{fish-speech-v1.4} claim to have trained their Fish Speech TTS using 20k Arabic audio hours, but they neither released the dataset nor described data collection. 

% Given how arabic is a low resource language, we need a way to curate a large enough dataset for training a more naturally sounding TTS system.

% Concatenative \cite{hunt1996unit} and HMM \cite{zen2007hmm} based speech synthesis paved the way for TTS in the early days. Curating a dataset would take a considerable amount of time \cite{halabi2016arabic}. Manual phonetic segmentation and alignment was costly and many approaches were taken to automate this process \cite{yuan2013automatic}. Even with automating alignment and segmentation, there is still the manual verification process where a team of linguists would be present to post-check the alignment. \citet{halabi2016arabic} has 3-4 hours, \citet{almosallam2013sassc} has around 7 hours recorded, corresponding to 51k words. Tacotron \cite{wang2017tacotron} alleviates the pain of having to curate a "purely sounding" dataset and opens the door to training on larger amounts of somewhat clean datasets. \cite{abdelali2022natiq} further extends this to Arabic. ClArTTS \cite{kulkarni2023clartts} creates a 12-hour single speaker corpus for classical and modern standard Arabic. More recently \cite{olamide2025arvoice} have introduced a substantially larger 83 hour hybrid dataset of which 10 hours are human and the rest is synthesized. 

% Given how arabic is a low resource language, we need a way to curate a large enough dataset for training a more naturally sounding TTS system.

\section{Data Collection}
Our target for data collection was to construct a high-quality Arabic speech dataset comprising audio segments from a diverse set of speakers, each ranging from 3 to 15 seconds in duration, accompanied by corresponding transcripts with and without diacritics. To collect the data, we tapped channels/groups on a popular social media platform where users share long-form audio content.  We stopped collecting after we aggregated roughly 20k raw audio hours.   We could have potentially collected significantly more data.  We processed the collected audio using the following steps:
% \begin{itemize}[leftmargin=*]
%     \item Voice Activity Detection (VAD):  We used the Silero open source VAD \cite{SileroVAD} to identify segments that contain human speech.  Silero was trained on audio from hundreds of languages and delivers SOTA VAD results while being lightweight.
%     \item Automatic Speech Recognition (ASR): We used the Fanar ASR model, which is conformer-based and was trained on roughly 15K audio hours of English and Arabic, including Modern Standard Arabic (MSA) and a variety of Arabic dialects \cite{team2025fanar}. The model reportedly delivers SOTA Arabic ASR results and is available via public APIs\footnote{\url{https://api.fanar.qa/docs}}.
%     \item Diacritization:  Since Arabic ASR output is typically undiacritized, we re-implemented the diacritizer described by \citet{elmallah2024arabic} with an in-house MSA corpus composed of 4 million words.  The model uses a morphologically informed character-based recurrent neural network model.  Our implementation has a word error rate (WER) of 5.5\% on the WikiNews corpus \cite{darwish2017arabic}.
%     \item Noise Filtering:  To filter out potentially noisy segments, we measured the noise level in the audio surrounding the audio segments that were extracted using VAD.  If the noise level was above -30 db in the 1 second leading or trailing an audio segment, we filtered that segment out. 
%     \item Length Filtering: We filtered out all segments shorter than 3 seconds and longer than 15 seconds.
% \end{itemize}
\vspace{-4pt}
    \paragraph{Voice Activity Detection (VAD):}  We used the Silero open source VAD \cite{SileroVAD} to identify segments that contain human speech.  Silero was trained on audio from hundreds of languages and delivers SOTA VAD results while being lightweight.
\vspace{-4pt}
    \paragraph{Automatic Speech Recognition (ASR):} We used the Fanar ASR model, which is conformer-based and was trained on roughly 15K audio hours of English and Arabic, including Modern Standard Arabic (MSA) and a variety of Arabic dialects \cite{team2025fanar}. The model reportedly delivers SOTA Arabic ASR results and is available via public APIs\footnote{\url{https://api.fanar.qa/docs}}.
\vspace{-4pt}
    \paragraph{Diacritization:}  We re-implemented the diacritizer described by \citet{elmallah2024arabic} with an in-house MSA corpus composed of 4 million words.  The model uses a morphologically informed character-based RNN model.  Our implementation has a word error rate (WER) of 5.5\% on the WikiNews corpus \cite{darwish2017arabic}.
\vspace{-4pt}
    \paragraph{Filtering:}  To filter out noisy segments, we measured the noise level in the audio surrounding the audio segments that were extracted using VAD.  If the noise level was above -30 db in the 1 second leading or trailing an audio segment, we filtered that segment out. We also filtered out all segments shorter than 3 seconds and longer than 15 seconds.

% \begin{figure}
%     \centering
%     \includegraphics[width=0.85\linewidth]{distribution.png}
%     \caption{Distribution of training data by segment length}
%     \label{fig:distribution}
% \end{figure}

After processing and filtering all the raw audio, we were left with roughly 4,000 hours of clean audio that is split into segments with corresponding transcription with and without diacritization.  %Figure \ref{fig:distribution} shows the length distribution of the audio segments.  
We randomly sampled 100 and 1,000 hours from the full datasets to test the effect of data size on TTS. We created two 100-hour subsets, with one maximizing the diversity of speakers and the other one minimizing it.  In all, we created 8 training sets: 4,000 hours with and without diacritics, 1,000 hours with and without diacritics, 100 hours with maximum and minimum speaker diversity with and without diacritics.

For evaluation, we constructed a test set containing audio segments from 59 different speakers.  For each speaker, we included 11 speech segments: one was used as an audio prompt for voice cloning, and the remaining ten were used to generate samples to evaluate the quality of the TTS using reference-based metrics.  To obtain audio samples for the speakers, we performed speaker diarization on a randomly selected subset of our collected data using Pyannote diarization model\footnote{\url{https://huggingface.co/pyannote/speaker-diarization-3.0}} \cite{Bredin23,Plaquet23}.  We then identified speakers who had more than 11 segments in a single audio file and spoke in MSA. The selected audio files, from which we extracted the test speakers, were excluded from the training data set to avoid data contamination.

\section{TTS Models}
To train our TTS models, We selected F5-TTS \cite{chen-etal-2024-f5tts}, a diffusion transformer model integrated with ConvNeXt V2, a fully convolutional masked autoencoder network with global response normalization.  F5-TTS delivers superior performance and enhanced efficiency with a real time factor of 0.15.  Given sufficiently large and diverse training data, the model is able to clone a voice with as little as a few seconds of a person's voice.  Once trained, the model is prompted with: 1) an audio segment for the voice to be cloned; 2) the transcript of the audio segment; and 3) the text to be vocalized.

 We trained our models with frame-wise batch size of 38,400, initial learning rate of 7.5e-5 and 20,000 warm-up steps. Our DiT is comprised of 22 layers of dimensionality of 1024 and 14 heads self-attention. Text inputs are projected to a 512-dimensional space with padding. We used Vocos \cite{siuzdak2023vocos} mel vocoder to synthesize audio output from predicted mel-spec features.  Given our 8 training sets, 8 different models were trained for 600k steps.

\section{Evaluation Metrics}
We used two reference-based evaluation metrics, namely:
%\begin{itemize}[leftmargin=*]
\vspace{-4pt}
    \paragraph{Word Error Rate (WER):}  We used the aforementioned Fanar ASR engine to transcribe the synthesized voice, and then we computed the WER of the transcript against the input text to the TTS model.  We used the jiwer Python library to compute WER\footnote{\url{https://github.com/jitsi/jiwer}}. This metric is more concerned with the correctness of the generated audio. 
\vspace{-4pt}
    \paragraph{SpeechBERTScore:} This measure is inspired by BERTScore. Reference and generated audios are encoded using speech self-supervised learning models, such as Wav2Vec, and the score is the cosine similarity between the encoded reference and generated audios \cite{saeki2024speechbertscore}.  Thus, it is more concerned with phonetic similarity between the reference and the generated audios. SpeechBERTScore has been shown to correlate well with human generated Mean Opinion Scores (MOS) \cite{saeki2024speechbertscore}.
    % \item SpeechBLEU: This is measure is akin to BLEU, which is used for measuring machine translation quality. To compute SpeechBLEU, reference and generated audios are encoded, as in SpeechBERTScore, the encoding vectors are clustered to create  discrete speech tokens, and the BLEU is computed over the generated speech tokens  \cite{saeki2024speechbertscore}. 
%\end{itemize}

\begin{table}[]
    \centering
    \begin{tabular}{c|c|c|c|c}
\multicolumn{3}{c}{Setup} & \multicolumn{2}{|c}{Results} \\ \hline
Diac.	&	Div.	&	Size	&	WER	&	SpeechBERT	\\ \hline
Yes	&	-	&	4,000	&	1.67	&	0.739	\\
Yes	&	-	&	1,000	&	2.55	&	0.729	\\
No	&	-	&	1,000	&	3.12	&	0.726	\\
No	&	-	&	4,000	&	3.18   &	0.732	\\
Yes	&	Max	&	100	&	3.48	&	0.719	\\
Yes	&	Min	&	100	&	3.62	&	0.712	\\
No	&	Max	&	100	&	4.06	&	0.704	\\
No	&	Min	&	100	&	5.20	&	0.706	\\
    \end{tabular}
    \caption{Results sorted (in ascending order) by WER}
    \label{tab:results}
\end{table}

\section{Experimental Results}
Table \ref{tab:results} lists the results for all models sorted by WER.  Also, we included some samples in an anonymized page here: \url{https://arabicnlp-t ts-paper-samples.great-site.net/}.  The results lead to a few notable conclusions, namely:
\vspace{-6pt}
\begin{itemize}[leftmargin=*]
    \item WER and SpeechBERTScore yield slightly different ordering of the different TTS models. The most notable difference concerns the non-diacritized 4,000 hour where WER ranks it lower than both 1,000 hour models while SpeechBERTScore ranks it higher than both of them.  This suggests that the non-diacritized 4,000 hour model yields higher phonetic similarity (better voice cloning) than the 1,000 hour models. %  In the first case, the difference in WER is small (0.06).  
    % Nonetheless, the difference between both metrics do not affect the overall conclusions.
    \item Diacritization leads to significantly better results compared to not using diacritics.  This consistent across all data sizes.
    % \item A model trained with diacritization with 1,000 hours beats a model trained on 4,000 hours but without diacritization with a significant margin. 
    \item Not surprisingly, increasing the number of training hours yields to improved TTS models.  The only exception to this was for the non-diacritized 4,000 and 1,000 models where the latter performs slightly better. Further investigation is required.
    \item Increasing the size of the training set yields higher phonetic similarity, as indicated by SpeechBERTScore across the board.  This suggests improved speaker voice (and environment) cloning.  This is clearly audible in the provided samples, where the 100 hour models are less able to capture the voice and environment properties of the reference compared to the 1,000 and 4,000 hour models.
    \item Maximum diversity for the 100 hour models produced better results than the minimum diversity.
\end{itemize}

\section{Conclusion}
The advancement of Arabic Text-to-Speech has been persistently hampered by two core challenges: the scarcity of large-scale public datasets and the inherent ambiguity of undiacritized text. In this work, we tackled these limitations by developing a robust, automated pipeline to curate a 4,000 hour Arabic speech corpus. Our pipeline effectively leverages existing technologies like VAD, ASR, and automatic diacritization to process raw, unannotated audio into a high-quality training resource.

While models trained with explicit diacritization consistently outperform their non-diacritized counterparts, we demonstrated that this performance gap narrows as the volume of training data increases, with for example the non-diacritized 1,000 hour model clearly beating the diacritized 100 hour models. This trend, observed across both WER and SpeechBERTScore metrics, suggests that with sufficient data, the model can implicitly learn the correct pronunciation from context, effectively compensating for the lack of explicit diacritics. This finding has profound implications, indicating that the high barrier to entry for creating quality Arabic TTS systems can be substantially lowered by focusing on large-scale data collection rather than perfect diacritization.  We will also release a high-quality, pre-trained Arabic TTS model capable of generating natural-sounding speech directly from undiacritized text. 

\section{Limitations and Future Work}
While our work presents a significant step forward in large-scale Arabic TTS, we acknowledge several limitations that offer avenues for future research.
\vspace{-4pt}
% \begin{itemize}[leftmargin=*]
    \paragraph{Quality Ceiling of the Automated Pipeline:} Dataset quality is bounded by the performance of the individual components like the ASR system and the automatic diacritizer. These errors propagate into the training data. Though our results suggest that the impact of these errors diminishes at a very large scale, they still represent a source of noise that could be reduced. %We could incorporate a human-in-the-loop verification step for a subset of the data or exploring more advanced, semi-supervised training methods to mitigate the impact of label noise.
    \vspace{-4pt}
    \paragraph{Reliance on Objective Metrics:} WER and SpeechBERTScore may not fully capture the perceptual qualities of synthesized speech, such as naturalness, prosody, emotional tone, and diacritization errors. Further, using the same ASR family for generating the training data and WER evaluation may introduce a favorable bias. Human evaluation is critical to validate real-world performance and user preference.
    \vspace{-4pt}
    \paragraph{Domain and Dialectal Scope:} % Our dataset was sourced from social media platforms and may be biased toward the collected genres. % like podcasts and vlogs. % Performance literary prose or highly formal speech, has not been explicitly tested. 
    Our study did not analyze the dialectal distribution of the final dataset or evaluate the TTS model's ability to render specific dialects. We only tested on MSA. Future research could focus on curating genre-specific and dialect-specific corpora to train more specialized and controllable Arabic TTS systems.
% \end{itemize}
% \begin{figure}[t]
%   \includegraphics[width=\columnwidth]{example-image-golden}
%   \caption{A figure with a caption that runs for more than one line.
%     Example image is usually available through the \texttt{mwe} package
%     without even mentioning it in the preamble.}
%   \label{fig:experiments}
% \end{figure}

% \begin{figure*}[t]
%   \includegraphics[width=0.48\linewidth]{example-image-a} \hfill
%   \includegraphics[width=0.48\linewidth]{example-image-b}
%   \caption {A minimal working example to demonstrate how to place
%     two images side-by-side.}
% \end{figure*}

% \begin{table*}
%   \centering
%   \begin{tabular}{lll}
%     \hline
%     \textbf{Output}           & \textbf{natbib command} & \textbf{ACL only command} \\
%     \hline
%     \citep{Gusfield:97}       & \verb|\citep|           &                           \\
%     \citealp{Gusfield:97}     & \verb|\citealp|         &                           \\
%     \citet{Gusfield:97}       & \verb|\citet|           &                           \\
%     \citeyearpar{Gusfield:97} & \verb|\citeyearpar|     &                           \\
%     \citeposs{Gusfield:97}    &                         & \verb|\citeposs|          \\
%     \hline
%   \end{tabular}
%   \caption{\label{citation-guide}
%     Citation commands supported by the style file.
%     The style is based on the natbib package and supports all natbib citation commands.
%     It also supports commands defined in previous ACL style files for compatibility.
%   }
% \end{table*}

\bibliography{custom}

% \appendix

% \section{Example Appendix}
% \label{sec:appendix}

% This is an appendix.

\end{document}